\def\year{2021}\relax
\documentclass[letterpaper]{article} 
\usepackage{aaai21}  
\usepackage{times}  
\usepackage{helvet} 
\usepackage{courier}  
\usepackage[hyphens]{url}  
\usepackage{graphicx} 
\usepackage{natbib}  
\usepackage{caption} 
\usepackage{amsmath}
\usepackage{amssymb}
\usepackage{subcaption}
\usepackage{color}
\title{SoFAr: Shortcut-based Fractal Architectures for Binary Convolutional Neural Networks}

\author{
    Zhu Baozhou\textsuperscript{\rm 1},
    Peter Hofstee\textsuperscript{\rm 1}\textsuperscript{\rm 2},
    Jinho Lee\textsuperscript{\rm 3},
     Zaid Al-Ars\textsuperscript{\rm 1}
    \\
}
\affiliations{

    \textsuperscript{\rm 1}Delft University of Technology, Delft, Netherlands \\
    \textsuperscript{\rm 2}
    IBM Research Austin, TX, USA\\
    \textsuperscript{\rm 3}
    Yonsei University, Seoul, Korea
}

\begin{document}

\maketitle

\begin{abstract}
Binary Convolutional Neural Networks (BCNNs) can significantly improve the efficiency of Deep Convolutional Neural Networks (DCNNs) for their deployment on resource-constrained platforms, such as mobile and embedded systems. However, the accuracy degradation of BCNNs is still considerable compared with their full precision counterpart, impeding their practical deployment. Because of the inevitable binarization error in the forward propagation and gradient mismatch problem in the backward propagation, it is nontrivial to train BCNNs to achieve satisfactory accuracy. To ease the difficulty of training, the shortcut-based BCNNs, such as residual connection-based Bi-real ResNet and dense connection-based BinaryDenseNet, introduce additional shortcuts in addition to the shortcuts already present in their full precision counterparts. Furthermore, fractal architectures have been also been used to improve the training process of full-precision DCNNs since the fractal structure triggers effects akin to deep supervision and lateral student-teacher information flow. Inspired by the shortcuts and fractal architectures, we propose two Shortcut-based Fractal Architectures (SoFAr) specifically designed for BCNNs: 1.~residual connection-based fractal architectures for binary ResNet, and 2.~dense connection-based fractal architectures for binary DenseNet. Our proposed SoFAr combines the adoption of shortcuts and the fractal architectures in one unified model, which is helpful in the training of BCNNs. Results show that our proposed SoFAr achieves better accuracy compared with shortcut-based BCNNs. Specifically, the Top-1 accuracy of our proposed RF-c4d8 ResNet37(41) and DRF-c2d2 DenseNet51(53) on ImageNet outperforms Bi-real ResNet18(64) and BinaryDenseNet51(32) by $3.29\%$ and $1.41\%$, respectively, with the same computational complexity overhead.
\end{abstract}

\section{Introduction}
Convolutional Neural Networks (CNNs) have become the paradigm of choice for visual recognition and made considerable breakthroughs in a wide range of visual tasks \cite{khan2020survey}, such as image recognition \cite{he2016deep,Verelst_2020_CVPR}, object detection \cite{liu2020deep}, and segmentation \cite{chen2017rethinking}. To practically deploy CNNs in the field, their efficiency has become a key differentiator, especially when targeting resource-limited embedded platforms. 

A significant amount research has been dedicated to increasing the efficiency of CNNs, including pruning \cite{He_2020_CVPR,wang2020apq}, quantization \cite{cai2020zeroq}, knowledge distillation \cite{romero2014fitnets,li2020block}, and efficient network design \cite{howard2019searching}. In low bit-width quantization, fixed-point integers are used instead of floating-point numbers~\cite{zhou2016dorefa}, where binarization is an extreme case of quantization. In BCNNs, the weights and/or activations are represented with only one bit, so the computation of the binary convolution can be completed by XNOR and Popcount bitwise operations \cite{rastegari2016xnor}. Binarization is the most efficient among the different bit-widths quantization methods, however, it results in accuracy degradation that is too large to be deployed in practice. 

The current methods to improve the accuracy of binarization can be divided into two categories \cite{zhuang2019structured}: value approximation and structure approximation. In value approximation, we preserve the topology of the full-precision CNNs during the binarization and seek a better local minimum for binarized weights/activations by either minimizing the quantization error \cite{mishra2017wrpn,shen2019searching,martinez2020training,bulat2019xnor,zhu2020lossless}, improving the loss function of the network \cite{wang2019learning,ding2019regularizing,mishra2017apprentice,hou2016loss}, or improving the quantization functions \cite{darabi2018bnn+,liu2018bi,lahoud2019self,yang2019quantization,qin2020forward}. In structure approximation \cite{liu2018bi,bethge2019binarydensenet,zhu2020nasb,zhuang2019structured}, the architecture of the binary CNNs is redesigned to approximate the original full-precision CNNs. The structure approximation focuses on the architecture design principles for efficient and accurate BCNNs, which is complementary to the value approximation. In this paper, our proposed method belongs to the structure approximation category.

Regarding structure approximation, Bi-real ResNet \cite{liu2018bi} and BinaryDenseNet \cite{bethge2019binarydensenet} show significant accuracy improvement without increasing the number of parameters, which indicates that adopting more shortcuts can help the training of BCNNs. In fractal architectures \cite{larsson2016fractalnet}, implicit deep supervision \cite{lee2015deeply} and student-teacher behavior \cite{ba2014deep} ensure the training of full-precision DCNNs without shortcuts, which shows the potential advantage of fractal architectures for dealing with the difficulty of training CNNs. Inspired by the shortcuts and fractal architectures, we propose our Shortcut-based Fractal Architectures (SoFAr) for BCNNs, which benefit from both the shortcuts and fractal architectures, and unifies them in one model.

The contribution of this paper is summarized as follows.
\begin{itemize}
\item We develop two shortcut-based fractal architectures (SoFAr) for BCNNs: the residual connection-based fractal architectures for binary ResNet and the dense connection-based fractal architectures for binary DenseNet.
\item Our proposed SoFAr improves on the accuracy of state-of-the-art shortcut-based BCNNs, achieving a better trade-off between efficiency and accuracy.
\item On classification tasks, diverse experiments are conducted to demonstrate the effectiveness of our proposed SoFAr.
\end{itemize}

\section{Related work}
In this section, we review and compare the recent work of compact architecture design and quantized CNNs.

\subsection{Compact architecture design}
Efficient architecture design has attracted lots of attention from researchers. $3 \times 3$ convolution has been replaced with $1 \times 1$ convolution in GoogLeNet \cite{szegedy2015going} and SqueezeNet \cite{iandola2016squeezenet} to reduce the computational complexity. Group convolution \cite{zhang2019differentiable}, depthwise separable convolution \cite{howard2019searching}, shuffle operation \cite{ma2018shufflenet}, and shift operation \cite{wu2018shift} are shown to reduce the computational complexity of traditional convolution. Instead of relying on human experts, neural architecture search techniques \cite{tan2019mnasnet,wu2019fbnet} can automatically provide optimized platform-specific architectures, achieving state-of-the-art efficiency.

\subsection{Quantized Convolutional Neural Networks}
Low bit-width quantization has been extensively explored in recent work, including reducing the gradient error \cite{gong2019differentiable}, improving the loss function of the network \cite{jung2019learning}, and minimizing the quantization error \cite{he2019simultaneously}. Moreover, mixed-precision quantized neural networks are developed to improve the performance further for low bit-width quantized neural networks. Using neural architecture search, mixed-precision neural networks \cite{wu2018mixed,li2020efficient,dong2019hawq} are developed to find the optimal bit-width (i.e., precision) for weights and activations of each layer efficiently.

Improving network loss function \cite{wang2019learning}, minimizing the quantization error \cite{martinez2020training}, and reducing the gradient error \cite{qin2020forward} have been studied to provide a better value approximation for BCNNs. Channel-wise Interaction based Binary Convolutional Neural Network (CI-BCNN) \cite{wang2019learning} uses a reinforcement learning model to mine the channel-wise interactions and impose channel-wise priors to alleviate the inconsistency of signs in binary feature maps. \cite{martinez2020training} obtain significant accuracy gains by minimizing the discrepancy between the output of the binary and the corresponding real-valued convolution. Information Retention Network (IR-Net) \cite{qin2020forward} is proposed to retain the information that consists of the forward activations and backward gradients. Regarding structure approximation, \cite{liu2018bi,bethge2019binarydensenet} adopts more shortcuts to help the training of BCNNs, which inspires aspects of our proposal.

\section{Binarization function}
In this section, we describe the binarization function that we adopt for our SoFAr, including the binarization of weights \cite{rastegari2016xnor} and activations \cite{liu2018bi}. We adopt the straight-through estimator (STE) \cite{bengio2013estimating} to approximate the gradient calculation for sign($\cdot$) function.

\subsection{Binarization of weights}
The forward propagation and backward propagation to binarize the weights are calculated as follows. $E$ and $L$ refer to the mean of the absolute value of the weights and the loss of the model, respectively. $W$ and $W_b$ represent the full precision weights and binary weights.
\begin{equation}
\begin{split}
\text{Forward: }{W_b} &= E  \times \text{sign}(W)  \\
\text{Backward: }\frac{{\partial L}}{{\partial W}} &= \frac{{\partial L}}{{\partial {W_b}}} \times \frac{{\partial {W_b}}}{{\partial W}} \approx  E \times \frac{{\partial L}}{{\partial {W_b}}} 
\end{split}
\end{equation}

\subsection{Binarization of activations}
The forward propagation and backward propagation to binarize the activations are calculated as follows. $A$ and $A_b$ represent the full precision activations and binary activations, respectively.
\begin{equation}
    \begin{split}
\text{Forward: }{A_b} &= \text{sign}(A) \\
\text{Backward: }\frac{{\partial L}}{{\partial A}} &= \frac{{\partial L}}{{\partial {A_b}}} \times \frac{{\partial {A_b}}}{{\partial A}} \\
\text{where}\frac{{\partial {A_b}}}{{\partial A}} &= \left\{ \begin{array}{l}
2 + 2A, -1 < A < 0\\
2 - 2A, 0 \le A < 1\\
0,\text{otherwise}
\end{array} \right.
    \end{split}
\end{equation}

\section{Shortcut-based fractal architectures}
In this section, we introduce our proposed SoFAr for BCNNs. Residual connections in ResNet \cite{he2016deep} and dense connections in DenseNet \cite{huang2017densely} are the most widely used shortcuts. Thus, we develop the residual connection-based fractal architectures for binary ResNet and the dense connection-based fractal architectures for binary DenseNet.

\begin{figure*}[h]
\centering
\includegraphics[width=0.8\textwidth]{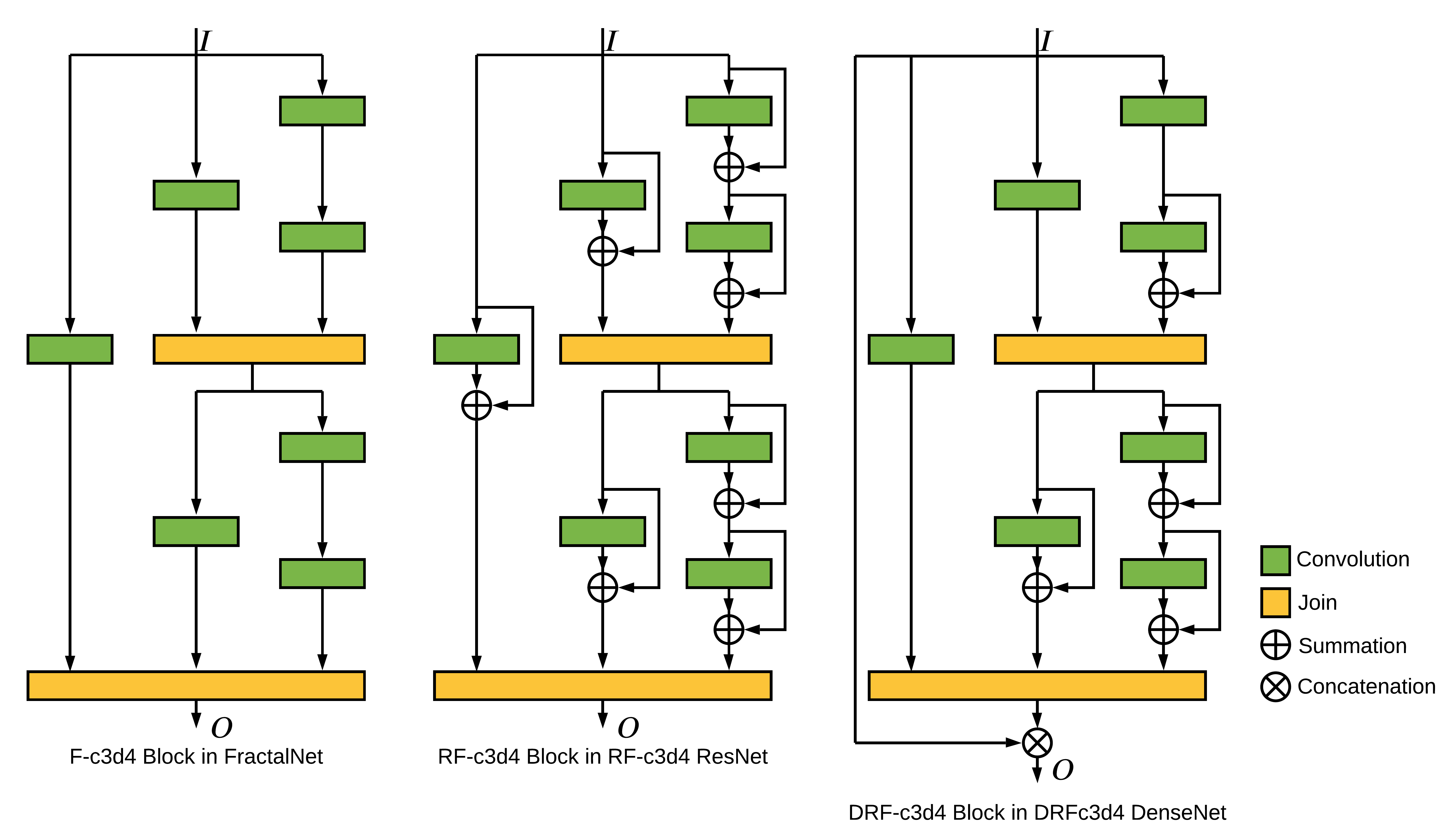}
\caption{The diagrams of fractal architectures, residual connection-based fractal architectures, and dense connection-based fractal architectures. Blocks in yellow and green refer to join and convolutional layers, respectively. $\oplus$ and $\otimes$ refer to feature aggregation of summation and concatenation, respectively.}
\label{fig1}
\end{figure*}

\begin{figure*}[h]
\centering
\includegraphics[width=0.8\textwidth]{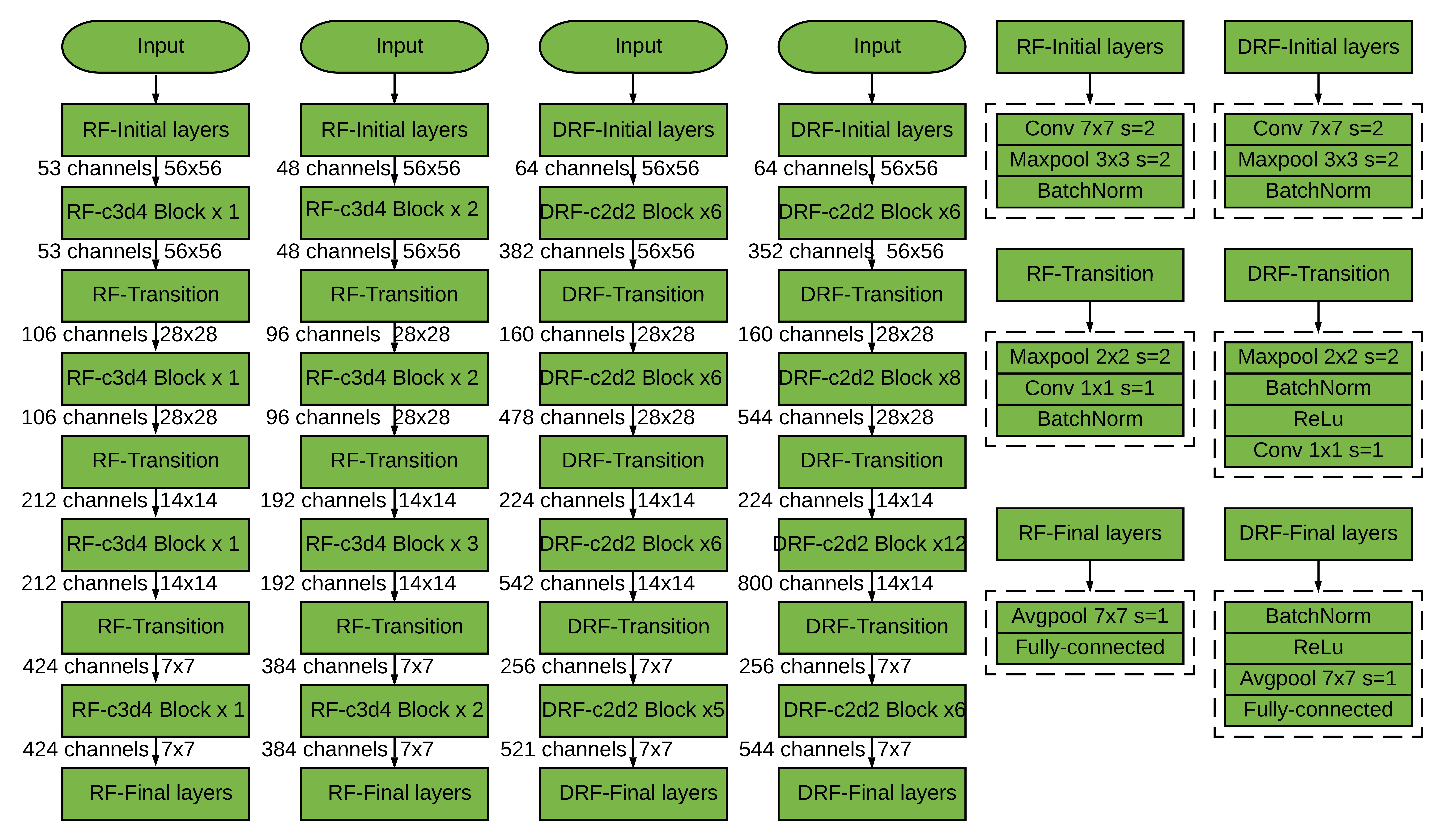}
\caption{The building blocks and an exemplary network structure of our shortcut-based fractal architectures.}
\label{fig4}
\end{figure*}

\subsection{Fractal architectures}
In fractal architectures for CNNs \cite{larsson2016fractalnet}, the truncated fractal with the index of $C$ can be defined as $F_C$, which can be a convolutional layer or convolutional block consisting of several convolutional layers. $I$ refers to the input activations of the truncated fractal. To have fractal architectures for CNNs, we have to define the base case and iteration rule. We use a single convolutional layer as the base case of the fractal architectures and define the successive fractals recursively as follows.
\begin{equation}
\begin{split}
&{F_1}(I) = \text{Conv}(I) \\
&{F_{C+1}}(I) = (F_C \circledast  F_C (I))  \odot  \text{Conv}(I)
\end{split}
\end{equation}

where $\circledast$ denotes composition operation and $\odot$ represents the join layer. The join layer is used to calculate the element-wise mean of all the inputs. It is worth noting that the neighboring join layers are collapsed into one single join layer as we expand the fractal architectures. The F-c3d4 block is shown to the left of Figure~\ref{fig1}, where the Batch Normalization and ReLU layers are omitted. $c3$ means that the number of columns of the fractal block is $c=3$. Similarly, $d4$ indicates that the longest depth between the input and output of the fractal block is $d=4$ convolutional layers. In a fractal block, we have $d=2^{c-1}$.

\begin{figure}[h]
\centering
\includegraphics[width=0.35\textwidth]{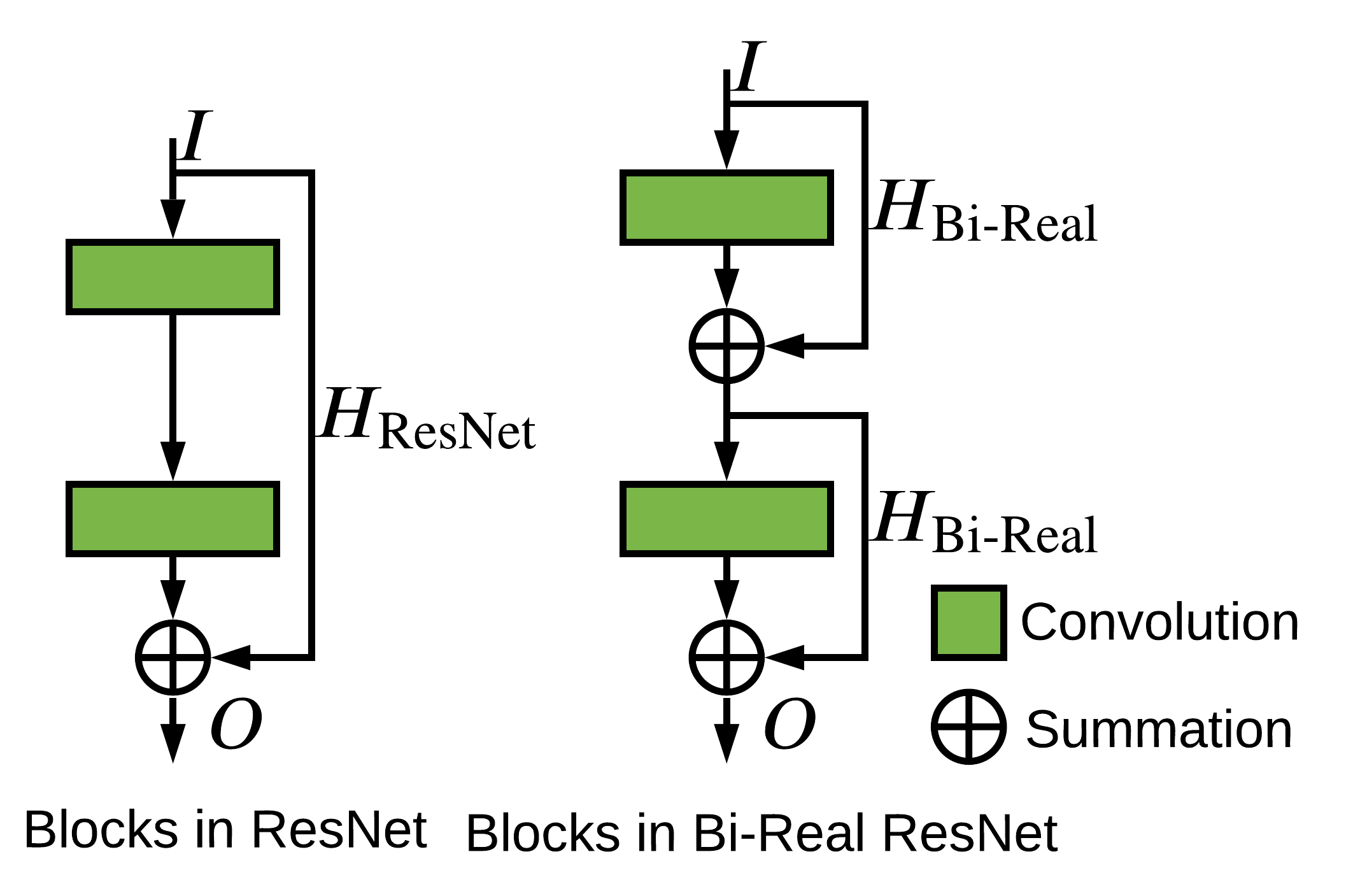}
\caption{Blocks in ResNet and Bi-real ResNet.}
\label{fig2}
\end{figure}

\subsection{Residual connection-based fractal architectures}
The residual connection can be expressed as follows where $\oplus$ refers to feature aggregation of summation. 
\begin{equation}
O = H(I) \oplus  I
\end{equation}

where $H$ is a nonlinear transform and includes convolutional layers, Batch Normalization layers, and ReLU layers.

In a block of ResNet as shown in Figure~\ref{fig2}, $H$ is composed of two convolutional layers.
\begin{equation}
{H_{\text{ResNet}}}(I) = \text{Conv} \circledast \text{Conv} (I) 
\end{equation}

In a block of Bi-Real ResNet, $H$ is one convolutional layer.
\begin{equation}
{H_{\text{Bi-Real}}}(I) = \text{Conv} (I) 
\end{equation}

In our residual connection-based fractal architectures, the RF-c3d4 block is shown to the middle of Figure~\ref{fig1}, where the number of columns and the longest depth is $c=3$ and $d=4$, respectively. All the convolutional layers are replaced with the convolutional layers with a residual connection compared with the fractal architectures of CNNs. We change the base case and the iteration rule for our residual connection-based fractal architectures. Specifically, the base case of the residual connection-based fractal architectures is as follows.
\begin{equation}
{F_{1}^{R}}(I) = \text{Conv}(I) \oplus  I
\end{equation}

Besides, we have successive fractals recursively as follows. 
\begin{equation}
{F_{C+1}^{R}}(I) = ( F_C^{R} \circledast   F_C^{R} (I) ) \odot ( \text{Conv}(I) \oplus  I )
\end{equation}

\subsubsection{Discussion} The fractal architectures and residual connection-based architectures facilitate the training of full-precision DCNNs since they share the key characteristic: large nominal network depth, but effectively shorter paths for gradient propagation during training. In Bi-Real ResNet, more residual connections, where the summation is used as the operation of feature aggregation, are introduced to help the training of BCNNs. Inspired by the residual connections and fractal architectures, our residual connection-based fractal architectures combine the advantages of fractal architectures and the adoption of residual connections in one unified model to resolve the difficulty of training BCNNs.

\begin{table*}[ht]
\begin{center}
\begin{tabular}{llllll}
\hline
Model  & Bit-width  & Top-1  & Top-5 & Parameters & Flops \\
\hline
Bi-real ResNet18(64)   & $b=32$   & $31.36\%$ & $11.57\%$  & -- & -- \\
Bi-Real ResNet18(64)  & $b=1$   & $40.42\%$ & $18.29\%$ & $33.18$Mbit & $1.64 \times 10^8$ \\
\hline
RF-c3d4 ResNet21(53)  & $b=32$   & $30.91\%$ & $10.94\%$   & -- & -- \\
RF-c3d4 ResNet21(53)  & $b=1$   & $37.58\%$ & $16.06\%$  & $32.63$Mbit & $1.46 \times 10^8$ \\
\hline
RF-c4d8 ResNet37(41)  & $b=32$   & $29.36\%$ & $10.31\%$   & -- & -- \\
RF-c4d8 ResNet37(41)  & $b=1$   & $\textbf{37.13}\%$ & $\textbf{15.63}\%$   & $32.24$Mbit & $1.28 \times 10^8$ \\
\hline
RF-c5d16 ResNet69(31)  & $b=32$   & $28.72\%$ & $9.88\%$   & -- & -- \\
RF-c5d16 ResNet69(31)  & $b=1$   & $37.66\%$ & $15.77\%$  & $32.16$Mbit & $1.14 \times 10^8$ \\
\hline
\hline
Bi-real ResNet34(64)   & $b=32$   & $29.24\%$ & $10.13\%$   & -- & -- \\
Bi-Real ResNet34(64)  & $b=1$   & $36.74\%$ & $15.36\%$  & $43.28$Mbit & $1.93 \times 10^8$ \\
\hline
RF-c3d4 ResNet41(48)  & $b=32$   & $27.94\%$ & $9.47\%$   & -- & -- \\
RF-c3d4 ResNet41(48)  & $b=1$   & $\textbf{35.62}\%$ & $\textbf{14.53}\%$   & $42.61$Mbit & $1.64 \times 10^8$ \\
\hline
RF-c4d8 ResNet77(35)  & $b=32$   & $27.61\%$ & $9.33\%$  & -- & -- \\
RF-c4d8 ResNet77(35)  & $b=1$   & $36.66\%$ & $15.07\%$   & $41.53$Mbit & $1.44 \times 10^8$ \\
\hline
\hline
BinaryDenseNet51(32)   & $b=1$   & $38.14\%$ & $16.80\%$  & $34.80$Mbit & $2.70 \times 10^8$ \\
\hline
DRF-c2d2 DenseNet51(53)  & $b=1$   & $\textbf{36.73}\%$ & $\textbf{15.54}\%$  & $34.53$Mbit & $2.97 \times 10^8$ \\
\hline
\hline
BinaryDenseNet69(32)   & $b=1$   & $36.26\%$ & $15.24\%$  & $41.95$Mbit & $2.82 \times 10^8$ \\
\hline
DRF-c2d2 DenseNet69(48)  & $b=1$   & $\textbf{35.20}\%$ & $\textbf{14.59}\%$  & $41.52$Mbit & $3.06 \times 10^8$ \\
\hline
\end{tabular}
\end{center}
\caption{Comparisons of binary ResNet and DenseNet variants on ImageNet.}
\label{table5}
\end{table*}

\subsection{Dense connection-based fractal architectures}

\begin{figure}[h]
\centering
\includegraphics[width=0.35\textwidth]{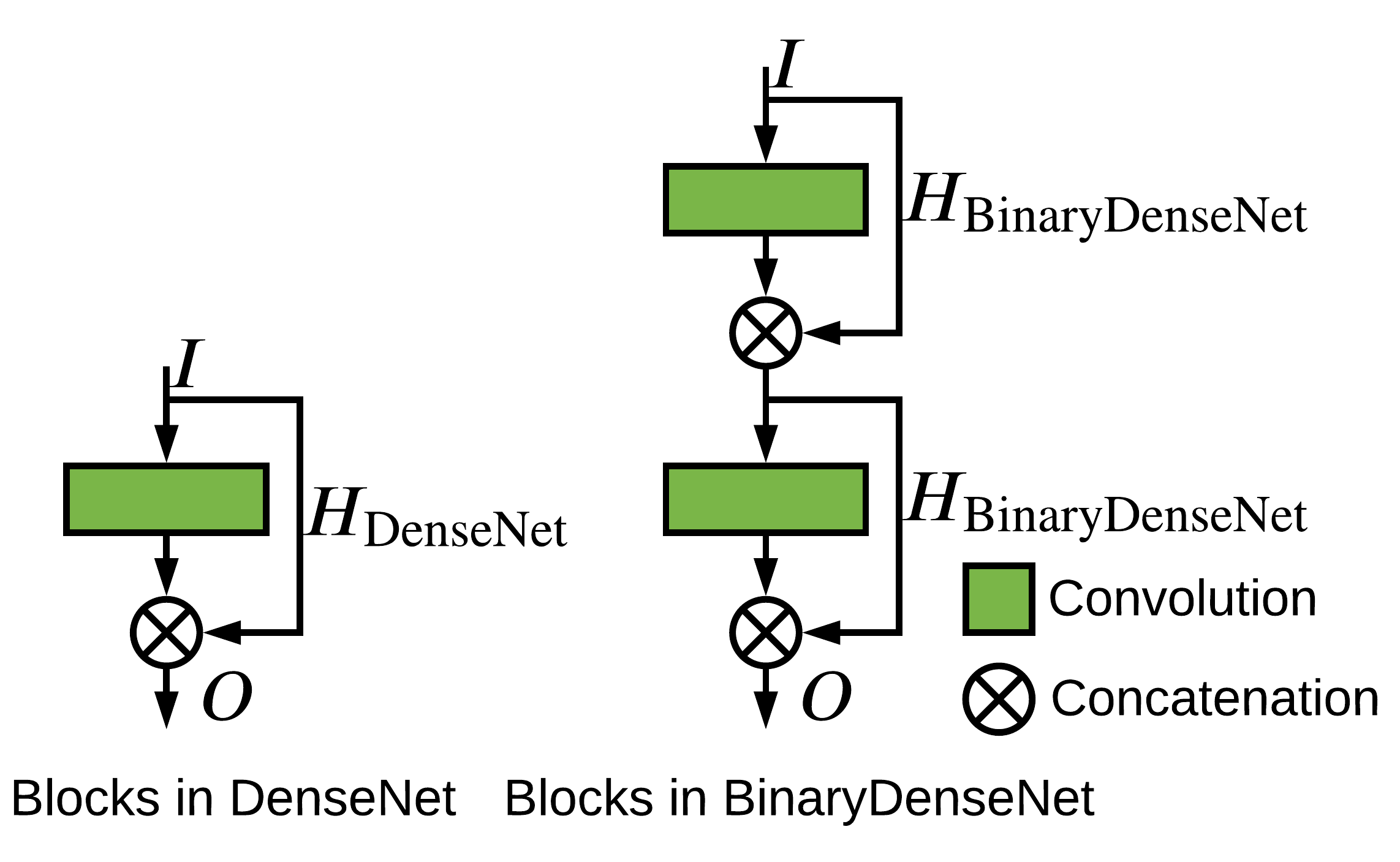}
\caption{Blocks in DenseNet and BinaryDenseNet.}
\label{fig3}
\end{figure}

The dense connection can be expressed as follows where $\otimes$ refers to feature aggregation of concatenation. 
\begin{equation}
O =  H(I) \otimes I
\end{equation}

where $H$ is a nonlinear transform and includes convolutional layers, Batch Normalization layers, and ReLU layers. In a block of DenseNet as shown in Figure~\ref{fig3}, $H$ is composed of two convolutional layers.
\begin{equation}
{H_{\text{DenseNet}}}(I) = \text{Conv} \circledast \text{Conv} (I) 
\end{equation}

In a block of BinaryDenseNet, $H$ is one convolutional layer. 
\begin{equation}
{H_{\text{BinaryDenseNet}}}(I) = \text{Conv} (I) 
\end{equation}

In our dense connection-based fractal architectures, the DRF-c3d4 block is shown to the right of Figure~\ref{fig1}, where the backbone, i.e., the convolutional and join layers, are the same as that in the fractal architectures. To introduce more shortcuts, our proposed DRF-c3d4 block is a combination of dense connection, residual connection, and fractal architectures. Two characteristics need to be clarified for our dense connection-based fractal architectures. In our DRF-c3d4 block, the fractal architectures are used to produce new features maps, which will concatenate with the feature maps of all preceding convolutional layers. In the fractal architectures of our DRF-c3d4 block, all the convolution layers, where the number of input channels is the same as the number of output channels, are associated with the residual connections. ${F_1}^{D}$ is the base case of our dense connection-based fractal architectures, where only one convolutional layer is used. ${F_1}^{D}$ is calculated as follows.
\begin{equation}
{F_1}^{D}(I) = \text{Conv}(I)
\end{equation}

We define the truncated fractal ${F_2}^{D}$ as follows where three convolutional layers are used and ${F_{1}^{R}}$ is the truncated fractal in our residual connection-based fractal architectures.
\begin{equation}
\begin{split}
&{F_2}^{D} = {F_1}^{D} \circledast {F_{1}^{R}}(I) \odot   \text{Conv}(I)\\
\end{split}
\end{equation}

We define the truncated fractal ${F_3}^{D}$ as follows where seven convolutional layers are used.
\begin{equation}
\begin{split}
&{F_3}^{D} =  {F_2}^{D} \circledast F_{2}^{R} (I) \odot \text{Conv}(I)
\end{split}
\end{equation}

At the end of our dense connection-based fractal architectures, we use feature aggregation of concatenation as follows.
\begin{equation}
\begin{split}
O={F_C}^{D} \otimes I
\end{split}
\end{equation}

\subsubsection{Discussion} Our proposed dense connection-based fractal architectures combine the advantage of the fractal architectures, residual connections, as well as dense connections in one unified model. Since the feature maps of all preceding convolutional blocks in DenseNet will be concatenated and reused, the fractal architectures are applied to produce new feature maps. Moreover, all the convolutional layers, where the number of input channels is the same as the number of output channels, adopt residual connection, so the shortcuts are used as often as possible. 

\section{Computational complexity}
We adopt the number of parameters as the metric for memory usage, and the number of Flops as the metric for computational efficiency. The number of parameters is measured as the summation of $32$bits times the number of floating-point parameters and $1$bit times the number of binary parameters in the model. The XNOR and Popcount bitwise operations can be executed by the current CPUs with a parallelism of $64$. Therefore, the Flops is calculated by the number of floating-point multiplications plus $1/64$ of the number of binary multiplication. To guarantee the fairness of the comparison, we scale the number of base channels of our SoFAr to match the computational complexity of the ResNet and DenseNet baselines.

As shown in Figure~\ref{fig4}, we describe our SoFAr with the input images of size $224 \times 224$. The left two columns are the residual connection-based fractal architectures, i.e., RF-c3d4 ResNet21(53) and RF-c3d4 ResNet41(48), respectively. $21$ and $41$ represent the depths of our residual connection-based fractal architectures, while $53$ and $48$ refer to their base number of channels, which are scaled to match the computational complexity of ResNet18 and ResNet34 after binarization, respectively. Similarly, we build DRF-c2d2 DenseNet51(53) and DRF-c2d2 DenseNet69(48) to compete with DenseNet51(32) and DenseNet69(32) after binarization \cite{bethge2019binarydensenet}, respectively. $51$ and $69$ refer to the depths of our dense connection-based fractal architectures, while $53$ and $48$ refer to the growth rate after scaling. We calculate the model depth with the criteria that every convolutional layer is recognized as one layer, which is different from that in \cite{bethge2019binarydensenet} (i.e., every block is recognized as a layer). To ensure consistency, binaryDenseNet28(64) and binaryDenseNet37(64) in \cite{bethge2019binarydensenet} are renamed as binaryDenseNet51(32) and binaryDenseNet69(32) in our paper.  The right two columns present the composition of the initial layers, transition block, and final layers in our SoFAr.

\begin{table*}[ht]
\begin{center}
\begin{tabular}{llllll}
\hline
Model  & Bit-width  & Top-1  & Top-5 & Parameters & Flops \\
\hline
Bi-real ResNet18(64)   & $b=32$   & $23.54\%$ & $6.55\%$  & -- & -- \\
Bi-Real ResNet18(64)  & $b=1$   & $28.48\%$ & $8.65\%$  & $18.18$Mbit & $1.67 \times 10^7$ \\
\hline
RF-c3d4 ResNet21(50)  & $b=32$   & $22.90\%$ & $5.90\%$  & -- & -- \\
RF-c3d4 ResNet21(50)  & $b=1$   & $\textbf{26.34}\%$ & $\textbf{7.89}\%$  & $18.07$Mbit & $1.53 \times 10^7$ \\
\hline
RF-c4d8 ResNet37(36)  & $b=32$   & $21.92\%$ & $5.87\%$  & -- & -- \\
RF-c4d8 ResNet37(36)  & $b=1$   & $26.67\%$ & $7.51\%$  & $17.57$Mbit & $1.42 \times 10^7$ \\
\hline
RF-c5d16 ResNet69(26)  & $b=32$   & $22.38\%$ & $5.98\%$  & -- & -- \\
RF-c5d16 ResNet69(26)  & $b=1$   & $26.85\%$ & $7.57\%$  & $17.63$Mbit & $1.38 \times 10^7$ \\
\hline
\hline
Bi-real ResNet34(64)   & $b=32$   & $21.71\%$ & $6.00\%$  & -- & -- \\
Bi-Real ResNet34(64)  & $b=1$   & $27.93\%$ & $8.37\%$  & $28.28$Mbit & $2.61 \times 10^7$ \\
\hline
RF-c3d4 ResNet41(45)  & $b=32$   & $22.03\%$ & $5.68\%$  & -- & -- \\
RF-c3d4 ResNet41(45)  & $b=1$   & $\textbf{25.36}\%$ & $\textbf{7.26}\%$  & $27.64$Mbit & $2.28 \times 10^7$ \\
\hline
RF-c4d8 ResNet77(32)  & $b=32$   & $21.77\%$ & $5.85\%$  & -- & -- \\
RF-c4d8 ResNet77(32)  & $b=1$   & $25.57\%$ & $6.86\%$  & $27.94$Mbit & $2.24 \times 10^7$ \\
\hline
RF-c5d16 ResNet149(22)  & $b=32$   & $22.47\%$ & $6.19\%$  & -- & -- \\
RF-c5d16 ResNet149(22)  & $b=1$   & $26.38\%$ & $7.83\%$  & $26.35$Mbit & $2.08 \times 10^7$ \\
\hline
\hline
BinaryDenseNet51(32)   & $b=1$   & $27.16\%$ & $7.27\%$ & $17.65$Mbit & $5.13 \times 10^7$ \\
\hline
DRF-c2d2 DenseNet51(48)  & $b=1$   & $\textbf{26.72}\%$ & $\textbf{7.51}\%$ & $17.51$Mbit & $5.32 \times 10^7$ \\
\hline
DRF-c3d4 DenseNet97(38)  & $b=1$   & $27.20\%$ & $7.74\%$ & $17.32$Mbit & $5.46 \times 10^7$ \\
\hline
\hline
BinaryDenseNet69(32)   & $b=1$   & $26.88\%$ & $7.52\%$ & $23.70$Mbit & $5.50 \times 10^7$ \\
\hline
DRF-c2d2 DenseNet69(44)  & $b=1$   & $\textbf{26.38}\%$ & $\textbf{7.32}\%$ & $23.33$Mbit & $5.67 \times 10^7$ \\
\hline
DRF-c3d4 DenseNet133(36)  & $b=1$   & $27.25\%$ & $7.68\%$ & $23.70$Mbit & $6.02 \times 10^7$ \\
\hline
\end{tabular}
\end{center}
\caption{Comparisons of ResNet and DenseNet variants on CIFAR-100.}
\label{table1}
\end{table*}

\section{Experimental results}
In this section, we evaluate our SoFAr on CIFAR-100 and ImageNet for the binarization of ResNet and DenseNet.

\subsection{Experimental results on ImageNet}
In this section, we present the experimental results of our SoFAr on ImageNet. Compared with both binary ResNet and binary DenseNet, our SoFAr shows significant accuracy improvement for BCNNs.
\subsubsection{ResNet variants on ImageNet}
For ResNet variants, we train a full precision model as an initialization for the BCNNs. During finetuning, the weights and activations are binarized, while the downsampling convolution layer or transition block remains in full precision in BCNNs. When training the full precision model, we reorder the layers from the order of "Conv-Bn-Relu" to the order of "Conv-Relu-Bn". Regarding the training settings and data processing for ResNet variants, we train a full precision model using a momentum optimizer and a weight decay of $1e-4$. We train $100$ epochs in total. The learning rate starts at $0.1$ and decays with a factor of $0.1$ at the step of $30$, $60$, and $90$. The Tanh function is inserted for the input activations of the convolution. During finetuning, we adopt an adam optimizer and a weight decay of $0.0$. We train $50$ epochs in total. The learning rate starts at $5e-4$ and decays at the step of $30$ and $40$. The Tanh function is replaced with the binarization function. We use a batch size of $256$.

As shown in Table~\ref{table5}, we present the experimental results of our residual connection-based fractal architectures on ImageNet. RF-c4d8 ResNet37(41) indicates that there are $4$ columns and $8$ convolutional layers on the longest path in a block of the residual connection-based fractal architectures. All the variants of our residual connection-based fractal architectures, including RF-c3d4 ResNet21(53), RF-c4d8 ResNet37(41), and RF-c5d16 ResNet69(31), achieve significant performance improvement compared with Bi-Real ResNet18. RF-c4d8 ResNet37(41) and RF-c3d4 ResNet41(48) improve the Top-1 accuracy by $3.29\%$ and $1.12\%$ compared with Bi-Real ResNet18(64) and Bi-Real ResNet34(64), respectively. Regarding the computational complexity, RF-c4d8 ResNet37(41) saves the number of parameters by $0.94$Mbit and the number of Flops by  $0.36 \times 10^8$ compared with Bi-Real ResNet18(64). Similarly, the number of parameters and the number of Flops required for our proposed RF-c3d4 ResNet41(48) are $0.67$Mbit and $0.29 \times 10^8$ less than those needed for Bi-Real ResNet34(64).

\subsubsection{DenseNet variants on ImageNet}
For DenseNet variants, we train from scratch for $100$ epochs with an adam optimizer and a weight decay of $0.0$. The learning rate starts at $0.002$ and decreases using a cosine annealing schedule until $0.0$. We use the method in \cite{glorot2010understanding} to initialize the weights. The Relu layer is removed from the "Bn-Relu-Conv" layers.

As shown in Table~\ref{table5}, we present the experimental results of our dense connection-based fractal architectures on ImageNet. The Top-1 accuracy of DRF-c2d2 DenseNet51(53) and DRF-c2d2 DenseNet69(48) are $1.41\%$ and $1.06\%$ better than those of BinaryDenseNet51(32) and BinaryDenseNet69(32), respectively. In terms of the computational overhead, DRF-c2d2 DenseNet51(53) and DRF-c2d2 DenseNet69(48) require $0.27 \times 10^8$ Flops and $0.24 \times 10^8$ Flops compared with BinaryDenseNet51(32) and BinaryDenseNet69(32), respectively, while they save the number of parameters by $0.37$Mbit and $0.37$Mbit, respectively. 

\subsection{Experimental results on CIFAR-100}
In this section, we present the experimental results of binary ResNet and DenseNet variants on CIFAR-100, which shows that our proposed SoFAr can improve the accuracy of binary ResNet and binary DenseNet with various depths.

\subsubsection{ResNet variants on CIFAR-100}

As shown in Table~\ref{table1}, we present the accuracy of residual connection-based fractal architectures for binarizing ResNet18 and ResNet34. All the variants of residual connection-based fractal architectures outperform Bi-Real ResNet baselines. Compared with Bi-Real ResNet18(64) and Bi-Real ResNet34(64), the Top-1 accuracy of our RF-c3d4 ResNet21(50) and RF-c3d4 ResNet41(45) are improved by $2.14\%$ and $2.57\%$, respectively. Considering the computational complexity, our RF-c3d4 ResNet21(50) use $0.11$Mbit and $0.14 \times 10^7$ Flops less than Bi-Real ResNet18(64). Our RF-c3d4 ResNet41(45) cost $0.64$Mbit and $0.33 \times 10^7$ Flops less than Bi-Real ResNet34(64). 

\subsubsection{DenseNet variants on CIFAR-100} 

As shown in Table~\ref{table1}, we present the accuracy of dense connection-based fractal architectures for binary DenseNet51(32) and DenseNet69(32). The Top-1 accuracy of our proposed DRF-c2d2 DenseNet51(48) and DRF-c2d2 DenseNet69(44) are $0.44\%$ and $0.50\%$ better than those of BinaryDenseNet51(32) and BinaryDenseNet69(32), respectively. The increased number of Flops for our proposed DRF-c2d2 DenseNet51(48) and DRF-c2d2 DenseNet69(44) are $0.19 \times 10^7$ and $0.17 \times 10^7$, respectively, while the decreased number of parameters for them are $0.06$Mbit and  $0.37$Mbit, respectively, compared with BinaryDenseNet51(32) and BinaryDenseNet69(32),
 
\subsection{Ablation study}
In the above section, we have shown the advantage of our SoFAr over the shortcut-based architectures for BCNNs, which indicate the benefits of the implicit student-teacher behavior and deep supervision of fractal architectures. In this section, we explore the role of shortcuts for our SoFAr.

\begin{table}[h]
\begin{center}
\begin{tabular}{@{}llll@{}}
\hline
Model  & Bit-width  & Top-1  & Top-5  \\
\hline
DRF-c2d2 DenseNet51(48)  & $b=1$   & $26.72\%$ & $7.51\%$ \\
DF-c2d2 DenseNet51(48)  & $b=1$   & $27.19\%$ & $7.26\%$ \\
\hline
DRF-c2d2 DenseNet69(44)  & $b=1$   & $26.38\%$ & $7.32\%$ \\
DF-c2d2 DenseNet69(44)  & $b=1$   & $26.63\%$ & $7.34\%$ \\
\hline
RF-c3d4 ResNet21(50)  & $b=1$   & $26.34\%$ & $7.89\%$ \\
F-c3d4 ResNet21(50)  & $b=1$   & $31.82\%$ & $9.70\%$ \\
\hline
RF-c3d4 ResNet41(45)  & $b=1$   & $25.36\%$ & $7.26\%$ \\
F-c3d4 ResNet41(45)  & $b=1$   & $40.14\%$ & $15.19\%$ \\
\hline
\end{tabular}
\end{center}
\caption{Ablation study results of CIFAR-100.}
\label{table7}
\end{table}

The architectures of DF-c2d2 DenseNet51(48) and F-c3d4 ResNet21(50) are obtained by removing all the residual connections from DRF-c2d2 DenseNet51(48) and RF-c3d4 ResNet21(50), respectively. As shown in Table~\ref{table7}, the residual connections can improve the Top-1 accuracy of DF-c2d2 DenseNet51(48) and DF-c2d2 DenseNet69(44) by $0.47\%$ and $0.25\%$, respectively. Similarly, the Top-1 accuracy degradation of F-c3d4 ResNet21(50) and F-c3d4 ResNet41(45) is $5.48\%$ and $14.78\%$ without residual connections.

\begin{table}[h]
\setlength{\tabcolsep}{3pt}
\small{
\begin{center}
\begin{tabular}{@{}llll@{}}
\hline
Model   & Top-1  & Top-5  \\
\hline
BNN ResNet18** \cite{courbariaux2016binarized}    & $57.80\%$ & $30.80\%$ \\
XNOR-Net ResNet18** \cite{rastegari2016xnor}      & $48.80\%$ & $26.80\%$ \\
TBN-ResNet18** \cite{wan2018tbn}                  & $44.40\%$ & $25.80\%$ \\
Trained Bin ResNet18** \cite{xu2019accurate}      & $45.80\%$ & $22.10\%$ \\
CI-Net ResNet18** \cite{wang2019learning}         & $43.30\%$ & $19.90\%$ \\
XNOR-Net++ ResNet18**                             & $42.90\%$ & $20.10\%$ \\
Bi-Real ResNet18 \cite{liu2018bi}                 & $43.60\%$ & $20.50\%$ \\
CI-Net ResNet18 \cite{wang2019learning}           & $40.10\%$ & $15.80\%$ \\
BinaryDenseNet51(32) \cite{bethge2019binarydensenet}  & $39.30\%$ & $17.60\%$ \\
Real-to-Bin ResNet18 \cite{martinez2020training}  & $\textbf{34.60}\%$ & $\textbf{13.80}\%$ \\
Bi-Real ResNet18(64)* \cite{liu2018bi}            & $40.42\%$ & $18.29\%$ \\
BinaryDenseNet51(32)* \cite{bethge2019binarydensenet}  & $38.14\%$ & $16.80\%$ \\
RF-c4d8 ResNet18(41)                              & $37.13\%$ & $15.63\%$ \\
DRF-c2d2 DenseNet51(53)  & $36.73\%$ & $15.54\%$  \\
\hline
TBN-ResNet34** \cite{wan2018tbn}                  & $41.80\%$ & $19.00\%$ \\
Bi-Real ResNet34 \cite{liu2018bi}                 & $37.80\%$ & $16.10\%$ \\
BinaryDenseNet69(32) \cite{bethge2019binarydensenet}  & $37.50\%$ & $16.10\%$ \\
Bi-Real ResNet34(64)* \cite{liu2018bi}            & $36.74\%$ & $15.36\%$ \\
BinaryDenseNet69(32)* \cite{bethge2019binarydensenet}  & $36.26\%$ & $15.24\%$ \\
RF-c3d4 ResNet41(48)   & $35.62\%$ & $14.53\%$ \\
DRF-c2d2 DenseNet69(48)  & $\textbf{35.20}\%$ & $\textbf{14.59}\%$  \\
\hline
Full-precision ResNet18 & $30.70\%$ & $10.80\%$ \\
Full-precision ResNet34 & $26.80\%$ & $8.60\%$ \\
\hline
\end{tabular}
\end{center}
\caption{Comparison with state-of-the-art methods on ImageNet. XNOR-Net++ ResNet18** is from \cite{bulat2019xnor}. * refers to the improved baseline reproduced in our paper. ** indicates the downsampling layers are binarized.}
\label{table8}
}
\end{table}

\subsection{Comparison to State-of-the-Art}
As shown in Table~\ref{table8}, we compare with state-of-the-art BCNNs on ImageNet. Except for the Real-to-Bin ResNet18 \cite{martinez2020training}, the Top-1 accuracy of our DRF-c2d2 DenseNet51(53) and DRF-c2d2 DenseNet69(48) achieve $36.73\%$ and $35.20\%$, respectively, and outperforms others by a large margin. More importantly, Real-to-Bin focuses on the minimization of quantization error between the BCNNs and their full precision counterparts, while our SoFAr work towards the architecture design for BCNNs. Thus, it is reasonable to expect that the performance of Real-to-Bin can be improved further when applying our proposed architectures.

\section{Conclusion}
In this paper, we proposed two shortcut-based fractal architectures for BCNNs: residual connection-based fractal architectures for binary ResNet, and dense connection-based fractal architectures for binary DenseNet. Benefiting from the fractal architectures and the adoption of shortcuts, our SoFAr can improve the performance of binary ResNet and binary DenseNet. Besides, we conduct experiments on classification tasks to show the advantage of our proposal. Under a given computational complexity budget, our proposed SoFAr achieves significantly better accuracy than current state-of-the-art BCNNs.

\clearpage
\bibliography{aaai21.bib}

\end{document}